\documentclass[conference]{IEEEtran}
\IEEEoverridecommandlockouts
\pdfoutput=1
\usepackage{times}
\usepackage{soul}
\usepackage{url}
\usepackage[hidelinks]{hyperref}
\usepackage[utf8]{inputenc}
\usepackage[small]{caption}
\usepackage{graphicx}
\usepackage{amsmath}
\usepackage{amsthm}
\usepackage{amssymb}
\usepackage{amsfonts}
\usepackage{booktabs}
\usepackage{bbding}
\usepackage{algorithm}
\usepackage{algorithmic}
\usepackage{bm}
\usepackage{multirow}
\usepackage{subfigure}
\title{Hippocampus-heuristic Character Recognition Network for Zero-shot Learning in Chinese Character Recognition}

\author{\IEEEauthorblockN{
		Shaowei Wang\IEEEauthorrefmark{1},	
		Guanjie Huang\IEEEauthorrefmark{2}, 
		Xiangyu Luo\IEEEauthorrefmark{3}
	}
	\IEEEauthorblockA{College of Computer Science and Technology, Huaqiao University, Xiamen, China\\
		Email: 
		\IEEEauthorrefmark{1}jem2rsc@gmail.com,
		\IEEEauthorrefmark{2}wbshgj@163.com,
		\IEEEauthorrefmark{3}luoxy@hqu.edu.cn \\
		Corresponding Author: Xiangyu Luo,     
		Email: luoxy@hqu.edu.cn }
}

\begin{document}

\maketitle

\begin{abstract}
The recognition of Chinese characters has always been a challenging task due to their huge variety and complex structures. The latest research proves that such an enormous character set can be decomposed into a collection of about 500 fundamental Chinese radicals, and based on which this problem can be solved effectively. While with the constant advent of novel Chinese characters, the number of basic radicals is also expanding. The current methods that entirely rely on existing radicals are not flexible for identifying these novel characters and fail to recognize these Chinese characters without learning all of their radicals in the training stage. To this end, this paper proposes a novel Hippocampus-heuristic Character Recognition Network (HCRN), which references the way of hippocampus thinking, and can recognize unseen Chinese characters (namely zero-shot learning) only by training part of radicals. More specifically, the network architecture of HCRN is a new pseudo-siamese network designed by us, which can learn features from pairs of input training character samples and use them to predict unseen Chinese characters. The experimental results show that HCRN is robust and effective. It can accurately predict about 16,330 unseen testing Chinese characters relied on only 500 trained Chinese characters. The recognition accuracy of HCRN outperforms the state-of-the-art Chinese radical recognition approach by 15\% (from 85.1\% to 99.9\%) for recognizing unseen Chinese characters.
\end{abstract}

\section{Introduction}

Traditional approaches to recognize Chinese characters require each character sample to be treated as a whole and labeled before training the model. As a result, the huge number of existing categories leads to so many parameters that need to be optimized. RAN \cite{b1} utilizes the method ~\cite{b10,b11} based on radicals learning and cleverly grasps the similarity among different characters.
\begin{figure}[htbp]
	\centerline{\includegraphics[width=0.9\columnwidth,height=0.2\textheight]{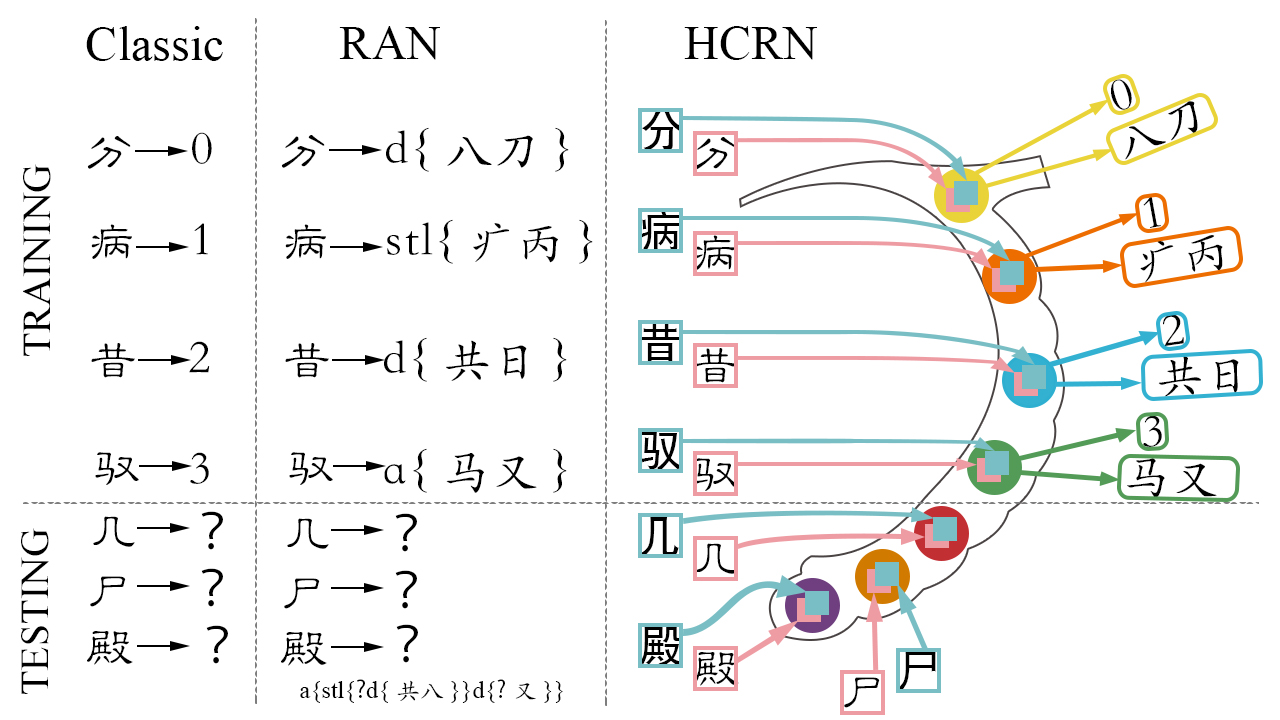}}
	\caption{The comparison among the traditional whole-characters based methods, radicals based RAN and proposed HCRN.}
	\label{fig1}
\end{figure}
Within this method, the size of categorizes is largely reduced in comparison to the conventional whole-character-based approaches, and the model has the capability of handling unseen or newly created characters. However, there are still some improvements in RAN: 1) Radical Equalization: It is unable to recognize the character that contains some radicals not learning from the training set; 2) Model Generalization: It only verifies the model's ability to recognize printed unseen characters and there is no public report on effectively applying this method to the recognition of more complex font styles.

Hippocampus-heuristic Character Recognition Network(HCRN) is inspired by the working mechanism of the hippocampus in the brain, and effectively solves the problem of radical equalization by taking advantage of the fact that people can memorize and summarize the character features, then use them to recognize other new characters, so the step of each radical must appear in the training stage doesn't matter. HCRN is improved based on the siamese network \cite{b2} and supplements a novel pseudo-siamese network with multiple losses. 
\iffalse
More importantly, HCRN  can be effectively applied to the task of recognizing more complex fonts. 
\fi
Fig.\ref{fig1} illustrates a clear comparison among traditional classified methods, RAN and HCRN on recognizing Chinese characters. In conventional methods~\cite{b9,b13}, each character is classified into a class category. If an unseen character class appears in the testing stage, it will be misclassified. In RAN, each character is decomposed into several radicals, and each radical will be regarded as a category. However, there exists the same problem that it is also unable to recognize some unseen radicals not appear in the training stage. While HCRN attempts to help the model achieve ``learn to learn" in a true sense like humans. For example, people encounter a new character, they will remember it and its radicals in a short time. When they encounter another new character next time, they will judge it according to their original prior knowledge and the recently learned knowledge. Such a learning way helps people master more Chinese characters as flexible as possible, which is also adopted in HCRN. As the example in the third column of Fig.\ref{fig1} shows, each Chinese character will be discomposed according to our own split strategy in the preprocessing stage, and each radical will act as an independent category. However, the difference between RAN and HCRN is that we don’t need to formulate the structure information of each Chinese character in the label. In the testing stage, when HCRN meets some unseen character, it will memorize that character and its radicals. These memories will be added to the prior knowledge for the next recognition task. Thus, HCRN has the ability to recognize unseen characters by only learning partial radicals.

To show its performance benefits and scalability, we evaluate it in three experiments. Surprisingly, HCRN has an excellent performance in each experiment. The main contributions of HCRN can be summarized as follows: 
\begin{itemize}
	\item HCRN can recognize unseen Chinese characters with only partial radicals, solving the problem of radical equalization.
	\item HCRN is robust and adapt for the recognition of Chinese characters with complex font styles or rotated angles.
	%\item We propose a new evaluation protocol for HCRN.
	\item We will announce the final radical splitting strategy used in the experiment for reference, which RAN didn't do that.
\end{itemize}

\section{Working Mechanism}

\subsection{Hippocampus-thinking}

The hippocampus of the brain is the main area that helps humans deal with long-term learning and memory events such as sound, light, and taste. According to some researches \cite{b12}, there are more than 100 billion neurons in the human brain, and each neuron stores a small piece of information(memories), similar to storing a character in a computer. And these neurons are connected to each other (nerve fibers), like a spider web. When we want to recall something, we essentially mobilize each neuron, let them gather information, and form a large information block through nerve fibers. These types of content like storing memories or searching for memories are exactly what the hippocampus does for us.

A more specific example about Chinese characters is as follows: When a learner meets a new Chinese character and its radicals, this information will be stored in a neuron as short-term memory. While encountering another new Chinese character, the hippocampus will
\begin{figure}[htbp]
	\centerline{\includegraphics[width=0.9\columnwidth,height=0.2\textheight]{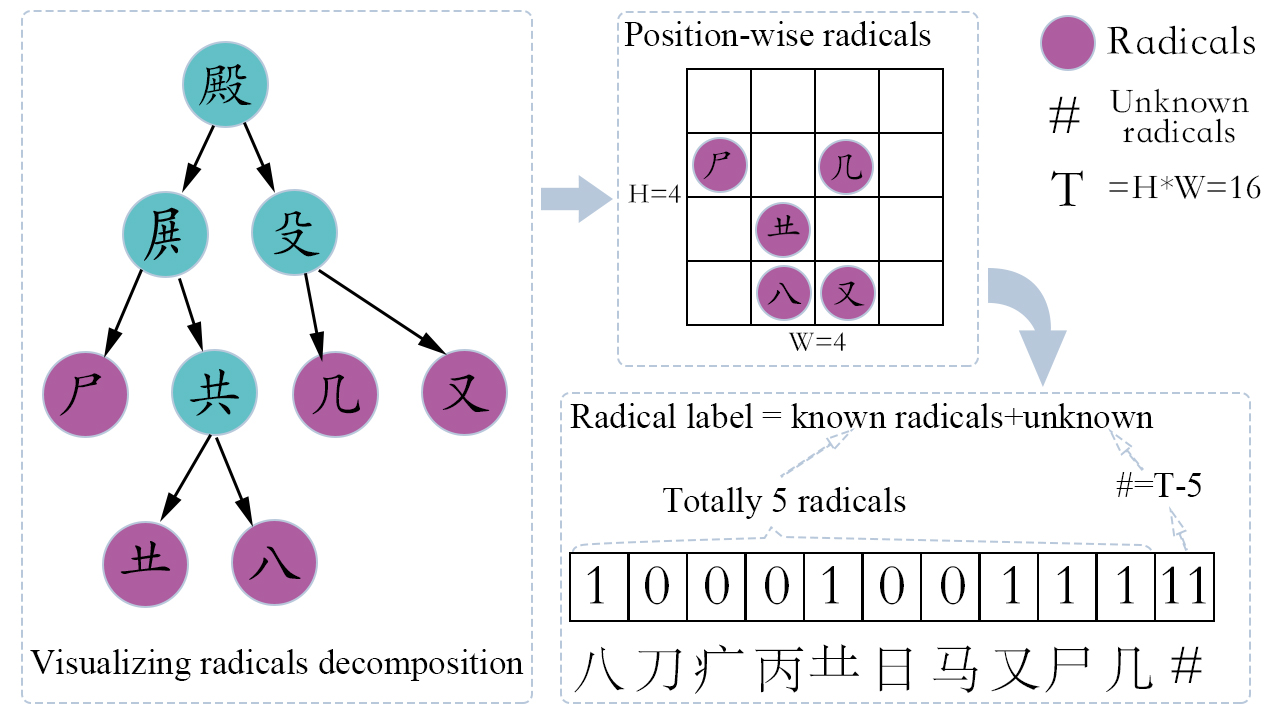}}
	\caption{The schematic diagram of HCRN's labeling strategies. }
	\label{fig2}
\end{figure}
help him find some connections (such as the same radicals) in the brain related to that character. The search scope includes the previously stored knowledge and short-term memory just stored. From this example, we can see that humans can learn new knowledge and use it at the same time. This performance in the computer can be described to give a model the ability to remember something new characteristics just learned in the testing stage and use them for later unseen character recognization in the same testing stage.

\subsection{No Structural Label Information}
In addition to the learning way of thinking like humans mentioned in the previous section, another highlight of HCRN is that we do not need to label the structural information of Chinese characters when we discompose them. RAN adopts an attention-based method and it needs to label the structural information. However, the method of attention mechanism is complicated, and it is time-consuming to mark structural information for each Chinese character. To avoid this unnecessary waste of resources and time, HCRN mainly adopts the ACE \cite{b4} method.

The ACE based model can exhibit a spatial distribution highly similar to the characters in the original text image after recognition. Fig.\ref{fig2} shows the process of tag generation in HCRN. The left part of the figure is an example of decomposing a character, the right-top part is one of the expected recognition results, and the rest right-bottom part refers to more details about the radical labels.

\section{Network Architecture}
\subsection{Backbone Network}
\begin{figure}[htbp]
	\centerline{\includegraphics[width=0.95\columnwidth,height=0.2\textheight]{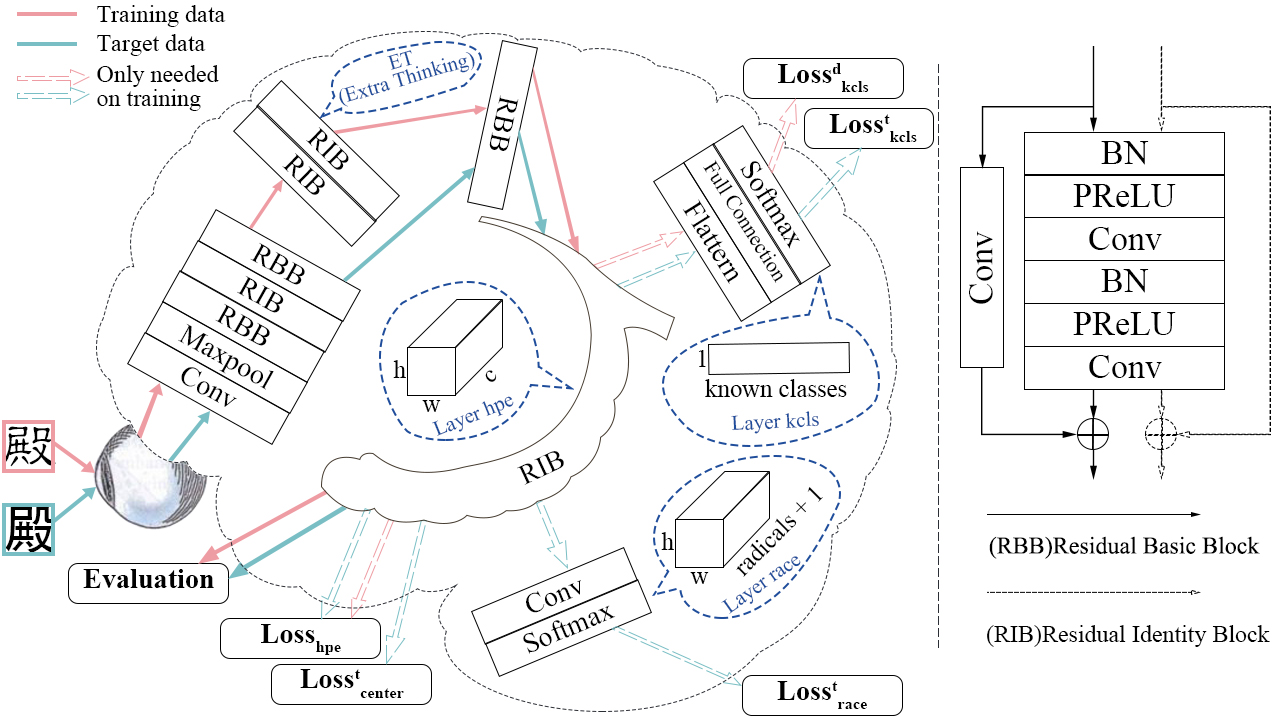}}
	\caption{The overall architecture of HCRN is a novel pseudo-siamese network with multiple loss functions. The pink and green lines are illustrations of ``training data" and ``target data", while the dash color lines represent what they only needed to process in the training stage. The right part of the figure is the cell of the backbone network, here we choose the basic block of ResNet V2 and make some improvements by replacing the original ReLU activation function with PReLU.}
	\label{fig3}
\end{figure}

The overall architecture of HCRN is illustrated in the left part of Fig.\ref{fig3}. We use the human brain as a schematic diagram to show the operating principle of HCRN. First, the model receives a pair of input samples which is composed of training data and target data. Next, the input samples will execute several processing steps. It is worth noting that the ``training data" has executed through one more step than the ``target data", which we call extra thinking (ET). Then the model will be measured through multiple loss functions we set, and finally applied to the verification task of text recognition.

As shown in the right part of Fig.\ref{fig3}, we choose ResNet V2 \cite{b7} as the main cell
of the backbone network in HCRN instead of ResNet V1 \cite{b15}. In terms of activation function, after many comparison experiments among ReLU, FReLU\cite{b14}, and PReLU\cite{b8}, PReLU was finally selected as the final configuration of the model.

\subsection{Multiple Loss Functions}
In HCRN, five independent loss functions form the final total loss function and more details will be presented in the following sections.

\subsubsection{Hippocampus Position-wise Embedding Loss Function.}
The loss function in this section represents the part of ``Layer hpe '' in Fig.\ref{fig3}. It mainly extracts the feature vector, position, and other information of the input character.

Assuming that the CNN layer extracts high-level visual representations denoted by a three-dimensional array of size \textit{H} × \textit{W} × \textit{C}. Then, the CNN output can be described as follows:
\begin{equation}
\resizebox{.8\linewidth}{!}{$
	\displaystyle
	\bm{A} =\bm{L_{hpe}}= \{ a_{111},..., a_{ijk} \}, A\in \mathbb{R}
	$}
\label{eq1}
\end{equation}
where $\textit{i} \in H, \textit{j}\in W, \textit{k}\in C $, and assume that the $\textit{L}_{hpe}$ of the input target data refers to $A^t$, the input training data refers to $A^d$, then the loss function can be summarized as follows:
\begin{equation}
\resizebox{.8\linewidth}{!}{$
	\displaystyle
	Loss_{hpe} = \sum_{i}^H \sum_{j}^W \sum_{k}^C \frac{1}{2}(a_{ijk}^d - a_{ijk}^t)^2 
	$}
    \label{eq2}
\end{equation}
where $\textit{a}_{ijk}^d$ denotes the feature vector of the input training data, and $\textit{a}_{ijk}^t$ denotes the input target data. Note that, here we use Mean Squared Error(MSE) to calculate the loss. Since the input data pair is two identical characters, which is essentially a regression task, so in such a scenario task, the MSE loss is a good loss function choice.  

\subsubsection{Known Character Classes Loss Function.}
This section of loss function represents the
part of ``Layer kcls'' in Fig.\ref{fig3}. The output K is presented by a sequence of one-hot encoded characters.
\begin{equation}
\resizebox{.8\linewidth}{!}{$
	\displaystyle
	\bm{K} =\bm{L_{kcls}}= \{ k_{1},..., k_{m} \}, k_{m}\in [0,1]
	$}
\end{equation}
where $\textit{K} = \textit{L}_{kcls} \in n \ vector$, \textit{n} is the number of known characters, and $\textit{k}_m$ refers to the probability of the \textit{m-}th character.

Assuming that the prediction of training data denotes $\textit{k}_{m}^d$, and target data denotes $\textit{k}_{m}^t$, then their relative label denotes $\textit{k}_{m}^{d^{*}}$ and $\textit{k}_{m}^{t^{*}}$.

The loss function of ``training data" can be summarized as follows:

\begin{equation}
\resizebox{.6\linewidth}{!}{$
	\displaystyle
	Loss_{kcls}^d = -\sum_{m}^n k_{m}^{d^{*}}  \log{k}_{m}^d
	$}
\end{equation}

Similarly, the loss function of ``target data" can be summarized as follows:
\begin{equation}
\resizebox{.6\linewidth}{!}{$
	\displaystyle
	Loss_{kcls}^t = -\sum_{m}^n k_{m}^{t^{*}} \log{k}_{m}^t
	$}
\end{equation}

Note that, this part is a traditional classified recognition task, so we utilize the cross-entropy loss function to calculate the loss. The performance in Fig.\ref{fig3} is we use a flatten layer and then connect with a full connection layer.

\subsubsection{Center Loss Function.}

Center Loss\cite{b3} is first proposed to improve the discriminative power of the deeply learned features. Intuitively, it can minimize the intra-class distance and maximize the inter-class distance. We apply the loss function to this paper, as formulated in Eq.\ref{eq6}.
\begin{equation}
\resizebox{.8\linewidth}{!}{$
	\displaystyle
	Loss_{center} = \sum_{i}^H \sum_{j}^W \sum_{k}^C \frac{1}{2}(a_{ijk}^t - C_{ijk}^t)^2 
	$}
    \label{eq6}
\end{equation}
where $\textit{a}_{ijk}^t$ refers to the target data mentioned in Eq.\ref{eq2}, $\textit{C}_{ijk}^t$ denotes the center of $\textit{a}_{ijk}^t$. The update of centers is consistent with the original center loss function, namely, we perform the update based on mini-batch instead of the entire dataset, and we also use a scalar $\alpha$ to control the learning rate of the centers.

Note that, we only calculate the center loss of target data, which is shown in Fig.\ref{fig3} (green dash line).

\subsubsection{Radical Aggregation Cross-Entropy Loss Function.}

In this section, we adopt the ACE method to calculate the loss of radicals. 
\iffalse
The advantage of ACE is that it can retain the spatial structure information of characters after recognition. In principle,  multiple radicals reorganize in spatial to form Chinese characters belongs to a 2D prediction problem. But luckily, ACE can easily handle this problem.
\fi
Assuming that the size of input is \textit{H} × \textit{W} × \textit{C}, then we use a 1 × 1 Conv layer to convert the channel from ``\textit{C}" to ``\textit{N}". The output can be represented as follows:
\begin{equation}
\resizebox{.8\linewidth}{!}{$
	\displaystyle
	\bm{R} =\bm{L_{race}}= \{ r_{111},..., r_{ijk} \}, R\in \mathbb{R}
	$}
\end{equation}
where $\textit{i} \in H, \textit{j}\in W, \textit{k}\in N $, namely the ouput shape of $\textit{R}$ is a three-dimensional array of size \textit{H} × \textit{W} × \textit{N}. \textit{N} is set to
\begin{table*}
	\centering
	\begin{tabular}{lccccccccccc}
		\toprule
		& 500 & 1000 & 2000 & 3000 &4000 & 5000 &6000 &7000 &8000 &9000 &100000\\
		\midrule
		RAN \  + Song   & -  & -  & 55.3  & 69.1 &74.4&77.3&78.6&82.5&83.3&84.2&85.1  \\
		HCRN + Song  & \bm{$93.4$}  & \bm{$97.8$}   & \bm{$99.1$}  & \bm{$99.1$} &\bm{$99.5$}&\bm{$99.6$}&\bm{$99.6$}&\bm{$99.7$}&\bm{$99.8$}&\bm{$99.8$}& \bm{$99.9$}    \\
		\midrule
		HCRN + Li   & 56.1  & 80.0  & 92.6  & 95.3 &95.9 &96.3&97.0&97.2&97.7&98.0&98.1   \\
		\bottomrule
	\end{tabular}
	\caption{Comparison of Accuracy Rate(AR \%) between RAN and HCRN with the same Song font style, and the performance of HCRN in recognizing complex Li font style.}
	\label{tab1}
\end{table*}
``radicals + 1", as is shown in ``Layer race" of Fig.\ref{fig3}, where ``radicals" refers to the number of total radicals we use in the training stage and the rest ``1" denotes the blank label to represent the unknown character. 

Note that, we only calculate the loss of target data, and assume that the prediction of the target data denotes $ r_{ijk}^t $, the accumulative radical probability of that target data denotes $ p_{k}^t $ ($ p_{k}^t $ = $\sum_{i=1}^{H}\sum_{j=1}^Wr_{ijk}^t$),which means the number of times that the \textit{k-th} radical occurs in the current sequence recognition. We set $\bm{Label^t}$ = $\{\bm{l_k^t}\}$, which means the true number of times that \textit{k-th} radical appears in the label.

We first normalize the accumulative probability of the \textit{k-}th radical $ p_{k}^t$ to $\bar{p}_{k}^t = p_{k}^t$ $/$ $T$, where $T = H$ × $W$. Then we normalize the relative labels $l_{k}^{t}$ to $\bar{l}_{k}^{t} = l_{k}^{t}$ $/$ $T$. Finally, the cross-entropy between $\bar{p}_{k}^t$ and $\bar{l}_{k}^{t}$ is expressed as:
\iffalse
Note that, we only calculate the loss of target data, and assume that the prediction of target data denotes $ r_{ijk}^t $, which means the number of times(accumulative probability) that the \textit{k-}th target data occurs in the current sequence recognition. We first normalize the accumulative probability of the \textit{k-}th character ($ r_{ijk}^t$), $\bar{r}_{ijk}^t = r_{ijk}^t$ $/$ $T$, where $T = H$ × $W$. Then we normalize the relative labels ($r_{ijk}^{t^{*}}$), $\bar{r}_{ijk}^{t^{*}} = r_{ijk}^{t^{*}}$ $/$ $T$. Finally, the cross-entropy between $\bar{r}_{ijk}^t$ and $\bar{r}_{ijk}^{t^{*}}$ is expressed as:
\fi

\iffalse
% here is the original content
Note that, we only calculate the loss of target data, and assume that the prediction of target data denotes $ r_{k}^t $, which means the number of times(accumulative probability) that the \textit{k-}th target data occurs in the current sequence recognition. The output 2D prediction $ r_{k}^t $ has height \textit{H} and width \textit{W}, so it can also be represented by $\sum_{h=1}^{H}\sum_{w=1}^{W}r_k^{hw^{t}}$, which denotes the prediction at the \textit{h-}th line and \textit{w-}th row. Now, we first normalize the accumulative probability of the \textit{k-}th character $ r_{k}^t$ to $\bar{r}_{k}^t = r_{k}^t$ $/$ $T$, where $T = H$ × $W$, and then we normalize the relative labels $r_{k}^{t^{*}}$ to $\bar{r}_{k}^{t^{*}} = r_{k}^{t^{*}}$ $/$ $T$. Finally, the cross-entropy between $\bar{r}_{k}^t$ and $\bar{r}_{k}^{t^{*}}$ is expressed as:
\fi

\begin{equation}
\resizebox{.8\linewidth}{!}{$
	\displaystyle
	Loss_{race} = -\sum_{k}^{N} \bar{l}_{k}^{t}  \log{\bar{p}}_{k}^t = -\sum_{k}^{N} \frac{l_{k}^{t}}{H * W}  \log{\frac{p_{k}^{t}}{H * W}} 
	$}
\end{equation}
\iffalse
\begin{equation}
\resizebox{.8\linewidth}{!}{$
	\displaystyle
	Loss_{race} = -\sum_{k}^{|C^{\varepsilon}|} \bar{r}_{ijk}^{t^{*}}  \log{\bar{r}}_{ijk}^t = -\sum_{k}^{|C^{\varepsilon}|} \frac{r_{ijk}^{t^{*}}}{H × W}  \log{\frac{r_{ijk}^{t}}{H × W}} 
	$}
\end{equation}
\fi
\iffalse
where $C^{\varepsilon} = C \cup \varepsilon$, with C being the total radical set and $\varepsilon$ the blank label, the length of $C^{\varepsilon} $ equals to $R_l$
\fi

Overall, the total loss of HCRN can be described as follows:
\begin{figure}[htbp]
	\centerline{\includegraphics[width=1\columnwidth ]{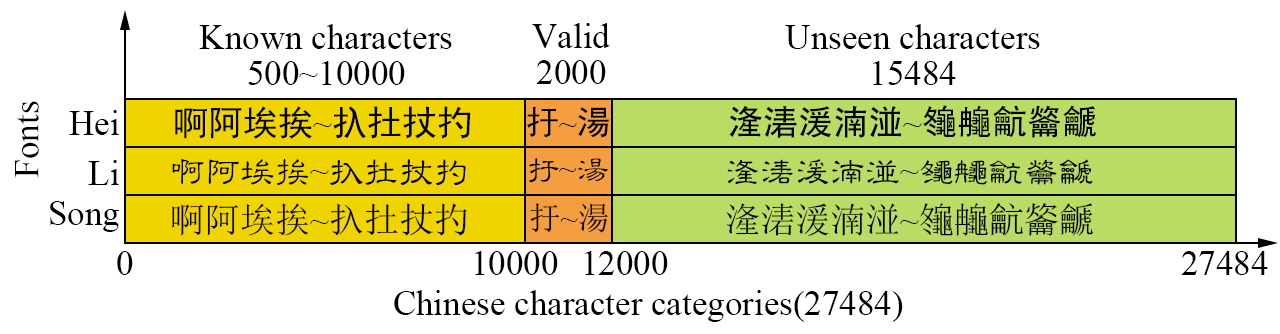}}
	\caption{Configuration visualization for the experiment of recognizing unseen Chinese characters. The ordinate axis represents the font style used in the experiment, and the abscissa axis represents the description of dividing into a training set, validation set, and testing set. }
	\label{fig4}
\end{figure}
\iffalse
% old equation
\begin{equation}
\resizebox{.9\linewidth}{!}{$
	\displaystyle
	Loss_{total} = Loss_{hpe} + Loss_{kcls}^d + Loss_{kcls}^t + \\ Loss_{center} + \\ 
    Loss_{race}$}
\end{equation}
\fi

\begin{align}
Loss_{total} = & Loss_{hpe} + 0.01 * Loss_{kcls}^d +   0.01 * Loss_{kcls}^t  \nonumber\\& + 0.003 * Loss_{center} +  Loss_{race}
\end{align}%
%where the hyperparameters we set for each loss function($Loss_{hpe}$, $Loss_{kcls}^d$, $Loss_{kcls}^t$, $Loss_{center}$, $Loss_{race}$) are (1, 0.01, 0.01, 0.003, 1)
where 0.01 and 0.003 denote the hyperparameters we set for the loss function($Loss_{kcls}$ and $Loss_{center}$), and the hyperparameters of other loss functions are set to 1.

\subsection{Inference for HCRN}
We evaluate the performance of HCRN on the embedding layer of $L_{hpe}$. As mentioned in Eq.\ref{eq1}, $L_{hpe}$ = $A$ = $\{a_{ijk}\}$. Assuming that $A^{test}$($\{a_{ijk}^{test}\}$) denotes the testing data. All the target data (means all used characters including training and testing data) we set as $H^t$, and $H^t$ = \{$A_c^{t}$\}= $\{a_{cijk}^t\}$, where c denotes the index of total characters. The evaluation method can be summarized as follows:
\begin{equation}
\resizebox{.6\linewidth}{!}{$
	\displaystyle
	D = \sum_{ijk} |a_{cijk}^t - a_{ijk}^{test}|
	$}
\label{eq10}
\end{equation}
\begin{equation}
\resizebox{.4\linewidth}{!}{$
	\displaystyle
	index = argmin(D)
	$}
\label{eq11}
\end{equation}
\begin{figure}[htbp]
	\centerline{\includegraphics[width=1\columnwidth, height=0.25\textheight]{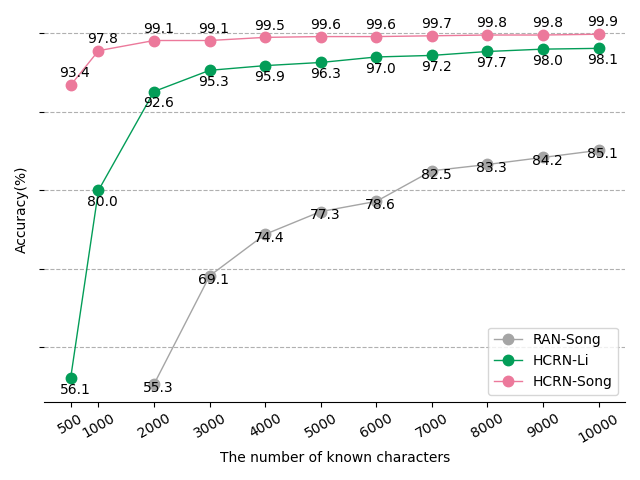}}
	\caption{The performance comparison among RAN and HCRN with the same Song font style, and HCRN with complex Li font style. }
	\label{fig5}
\end{figure}where D denotes the difference between the current target data and testing data, ``index" denotes the index of character which is the expected value of the model.

\section{Evaluation}
\iffalse
We evaluate HCRN by implementing it in four experiments, including recognition of unseen characters, recognition of seen characters, generalization research, and loss function research.
\fi

\subsection{Implementation Details}
\subsubsection{Training Scene.}
All experiments are implemented on an Ubuntu workstation with a 24-core Intel CPU, 4 NVIDIA TITAN RTX GPU, and 64G RAM.

\subsubsection{Dataset and Training Strategy.}
All the Chinese characters we use as datasets are taken from Unicode and generated with different font styles. In terms of the input data of the model, we use Hei font style as the font of ``target data", which is fixed, and Song font style as the font of ``training data", which is variable, because we also verify more complex font styles in subsequent experiments.
\iffalse
\subsubsection{Dataset and Training Strategy.}
We employ ResNet V2 architecture as our backbone network and replace the original ReLu activation function with PReLU for the best effect. In ResNet V2, the number of output channels in each block is (64, 128, 256, 512). The number of output channels of each convolutional layer in the same block remains unchanged. In terms of the input data of the model, we use Hei font style as the font of ``target data", which is fixed, and Song font style as the font of ``training data", which is variable, because we also verify more complex font styles in subsequent experiments.
\fi

\subsection{Results on Recognizing Unseen Characters}
We will verify the effectiveness of HCRN on recognizing unseen Chinese characters by comparing RAN in this section.

In the experiment, for a fair comparison, we choose 27484 Chinese characters with Sont font style like RAN does that, but the difference is we only use 299 radicals to represent the whole character set, while RAN has used 361 radicals and more than 30 structures. As is shown in Fig.\ref{fig4}, we divide the character set into a training set, validation set, and testing set. We gradually increase the training set
\begin{table*}
	\centering
	\begin{tabular}{lccccccccccc}
		\toprule
		&0-shot& 1-shot & 2-shot & 3-shot & 4-shot & 5-shot & 6-shot & 10-shot & 14-shot & 18-shot & 22-shot\\
		\midrule
		Zhong  &-& 4.0  & 16.3  & 43.7  & 62.4 & 73.1 & 74 &84.7 &87.9 &89.7 & 91.3   \\
		VGG14  & - & 23.4  & 58.4  & 74.3 & 82.2 & 83.1 & 86.4 & 90.3 & 90.7 & 92.9 & 95.4   \\
		RAN    & - & 80.7  & 85.7  &  87.9 & 88.3  & 89.9 & 91.3 & 94.1 & 94.6 & 95.6 & 96.2\\
		HCRN   & \bm{$95.6$}  & \bm{$94.4$}  &  \bm{$95.0$}  & \bm{$95.4$} & \bm{$95.3$} & \bm{$95.4$} & \bm{$95.3$} & \bm{$96.0$} & \bm{$96.0$} & \bm{$96.2$} &\bm{$96.9$} \\
		
		\bottomrule
	\end{tabular}
	\caption{Comparison of Accuracy Rate(AR \%) among traditional whole-characters based approaches(Zhong, VGG14), RAN and HCRN.}
	\label{tab2}
\end{table*}
from 500 to 10000 Chinese characters. We choose 2000 characters as the validation set and 15,484 characters as the testing set, which is consistent with RAN. Note that, due to the particularity of our model, in addition to using Song font as the font style of the ``train data", we also use Hei font as the font style of the ``target data".

The performance of HCRN is illustrated in Tab.\ref{tab1}, where only 500 training Chinese characters can successfully recognize about 93.4\% unseen 15,484 Chinese characters. However, the best accuracy of RAN by using 10,000 training Chinese characters can only achieve 85.1\%. Surprisingly, the recognition rate of HCRN is directly
\begin{figure}[htbp]
	\centerline{\includegraphics[width=1\columnwidth ]{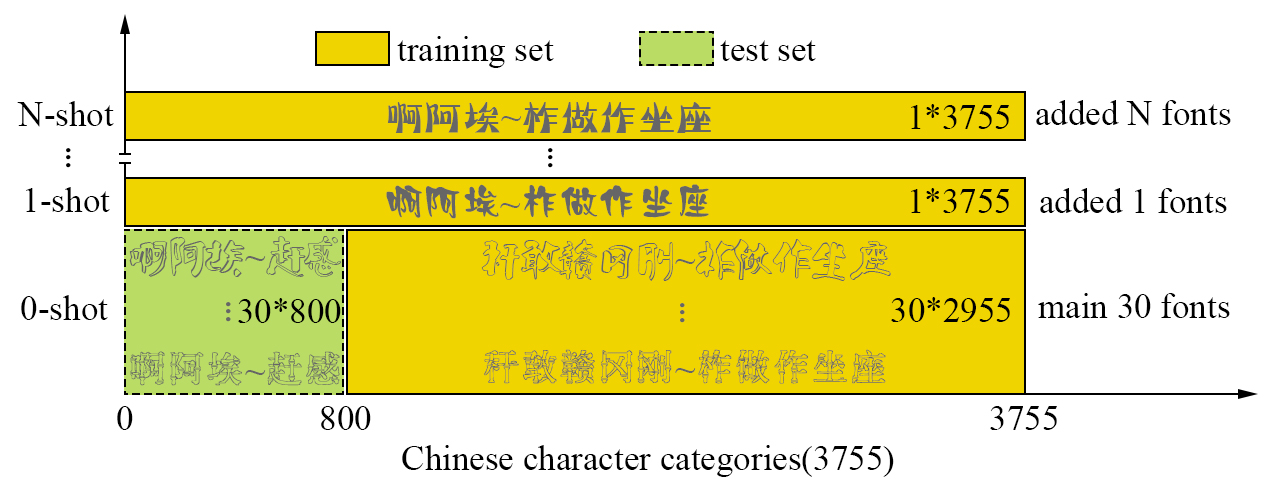}}
	\caption{Configuration visualization  for the experiment of recognizing seen Chinese characters.}
	\label{fig6}
\end{figure}
proportional to the growth of the training data. Finally, with 10,000 training characters used, the accuracy of the model has achieved 99.9\%. A more notable comparison is shown in the Fig.\ref{fig5}. The pink line represents HCRN and the grey line represents RAN.

We also conduct an experiment with a complex font style to see the recognition ability of HCRN. As is shown in Fig.\ref{fig4}, we adopt Li font to evaluate the model. The performance of HCRN is illustrated in Tab.\ref{tab1}, where 500 characters with Li font style can successfully recognize about 56.1\% 15,484 unseen Chinese characters. We can get the conclusion that it is more difficult to recognize Chinese characters with complex font styles. However, the overall performance of HCRN is still better than RAN. A clearer comparison is shown in Fig.\ref{fig5}, where the green line represents HCRN, and the grey line represents RAN.

\subsection{Results on Recognizing seen Characters}

\begin{figure}[ht]
	\centering
	\subfigure[main 30 fonts]{
		\label{Fig.sub.1}
		\includegraphics[width=0.42\columnwidth ]{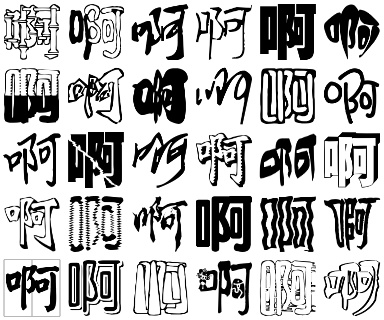}}
	\subfigure[added 22 fonts]{
		\label{Fig.sub.2}
		\includegraphics[width=0.42\columnwidth]{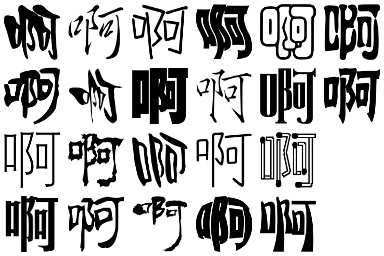}}
	\caption {The overall font styles used in the experiment of recognizing seen Chinese characters.}
	\label{fig7}
\end{figure}

We have verified the effectiveness of HCRN on recognizing seen Chinese characters by comparing RAN and the traditional method in this section. As is shown in Fig.\ref{fig6}, we divide the dataset into a testing set with 30(fonts) * 800 characters and a training set with 30 * 2955 characters. Additionally, we use 3755 characters with other 22 font styles as the second part of the training set like RAN does that. Similarly, when we add 3755 characters with other N font styles as the second training set, we call this experiment N-shot. In the experiment, we increase the extra font styles from 1 to 22. The overall font styles we have adopted are visualized in Fig.\ref{fig7}.

As is illustrated in Tab.\ref{tab2}, ``Zhong" is the proposed method in \cite{b5}, ``VGG14" is the improvement of ``Zhong", where the CNN architecture is replaced with VGG 14\cite{b6} and other parts keep unchanged. We can easily get the conclusion that HCRN achieves the best performance at each shot. Additionally, we also verify the performance of HCRN at 0 shot. A clearer comparison is declared in Fig.\ref{fig8}, where the pink line denotes HCRN, and it performs best among these comparison methods.

\subsection{Results on Generalization Research}

In this section, we will evaluate the value of Extra Thinking(ET) and the performance of HCRN to recognize the rotated Chinese characters.

As is shown in Fig.\ref{fig3}, Extra Thinking(ET) is a module in HCRN. It is worth 
noting that if we remove the module of ET, the whole architecture of HCRN degenerates into a siamese network.
\iffalse
\begin{table}[htbp]
	\centering
	\resizebox{.75\linewidth}{!}{
	\begin{tabular}{|l|c|c|}
		\hline
		&$\checkmark$ with ET & $\times$ without ET \\
		\hline
		Accuracy  &99.14  & 99.08  \\
		\hline
	\end{tabular}}
	\caption{The performance comparison between HCRN with ET and without ET.}
	\label{tab3}
\end{table}
\fi
\begin{table}[htbp] \scriptsize
	\centering
	
	\begin{tabular}{|c|c|c|c|c|c|c|}
		\hline
		\multirow{2} * {GB(0.5)}& \multicolumn{3}{|c|}{Random Rotation} & \multicolumn{3}{|c|}{Accuracy}  %\multirow{2} * {Diff} \\
		\\
		\cline{2-7}
		~&[-15,15] & [-30,30] &[-45,45]& $\checkmark$ ET & $\times$ ET &Diff \\
		\hline
		~ & ~ &~ & ~ &99.14 & 99.08 & 0.06\\
		\hline
		$\checkmark$ & ~ &~ & ~ &99.41 & 99.35 & 0.06\\
		\hline
		$\checkmark$ & $\checkmark$ &~ & ~ &98.92 & 98.81 & 0.11\\
		\hline
		$\checkmark$ & ~ &$\checkmark$ & ~ &97.74 & 96.66 & 1.08\\
		\hline
		$\checkmark$ & ~ &~ & $\checkmark$ &95.53 & 91.64 & 3.89\\
		\hline
	\end{tabular}
	\caption{The performance evaluation of ET's effect in HCRN and HCRN's generalization ability by adding Gaussian Blur(GB) and random rotation for characters.}
	\label{tab4}
\end{table}

In this part of the experiment, we use 3755 Chinese characters with Li font style, then we divide them into 1855 characters of the training set and 1900 characters of the testing set. From the first row of tab.\ref{tab4}, we can get the conclusion that HCRN with ET can recognize about 99.14\% unseen 1900 characters and about 99.08\% of accuracy without ET.
Although ET improves the performance of HCRN, the improvement is a little weak. Thus, we make some more experiments on recognizing rotated Chinese characters to find out the value of ET and the scalability of HCRN. More details are illustrated in Tab.\ref{tab4}. We add the Gaussian blur with the coefficient of 0.5 and random rotation angles to the characters. We can draw the following conclusions from the chart:
\begin{figure}[ht]
	\centerline{\includegraphics[width=1\columnwidth, height=0.25\textheight]{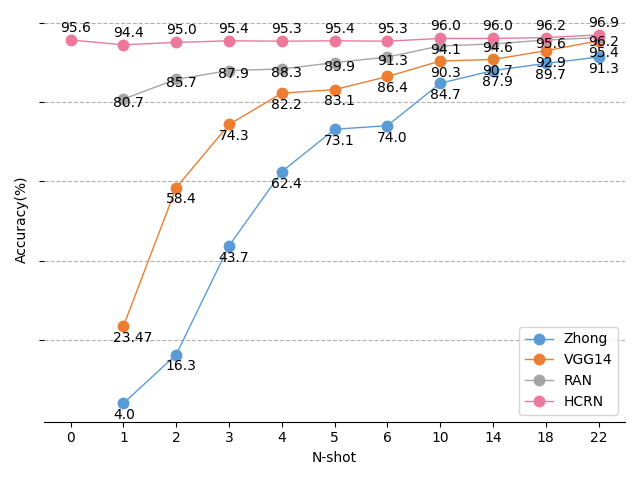}}
	\caption{The performance comparison among Zhong, VGG14, RAN and HCRN with respect to the number of newly added training fonts N.}
	\label{fig8}
\end{figure}
\begin{itemize}
	\item Although ET has little effect on original recognition tasks, with added blur and increased rotated angle, the difference in recognition accuracy between ET and without ET reaches 3.89.
	\item HCRN still performs excellently on recognizing rotated Chinese characters. While with 1855 characters for training, it can successfully identify about 95.53 \% unseen 1900 Chinese characters.
\end{itemize}

\subsection{Results on Loss Function Research}

In this section, we verify the impact of four loss functions on the HCRN recognition results. The experimental configuration we use is the same as the verification experiment on ET. More details are illustrated in Tab.\ref{tab5} and the observations are as follows:
\begin{itemize}
	\item From row 1 to 4 in the table, we see that $loss_{kcls}$ performs best in the case of a single loss function.
	\item From row 5 to 10 in the table, we see that the combination of $loss_{hpe}$ and $loss_{race}$ performs best in the case of pairwise loss function combination.
	\item From row 11 to 14 in the table, we see that the combination of $loss_{hpe}$, $loss_{kcls}$ and $loss_{race}$ performs best in the case of three pairs of the loss function.
	\item Row 8, 12, and 13 compare the impact of adding other loss functions on recognition accuracy under the premise of fixing the two loss functions of
	\begin{table}\small
		\centering
		\begin{tabular}{|c|c|c|c|c|c|c|c|c|}
			\hline
			& $loss_{hpe}$ & $loss_{kcls}$  &$loss_{center}$  & $loss_{race}$  & $acc$  \\
			\hline
			1  &   &   &  & $\checkmark$ & 29.0  \\
			\hline
			2  &   &   & $\checkmark$  &  & 12.99 \\
			\hline
			3 &   & $\checkmark$  &   &  & 40.38  \\
			\hline
			4 & $\checkmark$  &   &   &  & 4.37  \\
			\hline
			5 &   &   & $\checkmark$  &$\checkmark$ &31.91  \\
			\hline
			6 &   & $\checkmark$   &  &$\checkmark$ &35.96  \\
			\hline
			7 &   & $\checkmark$   & $\checkmark$  & &46.42  \\
			\hline
			8 & $\checkmark$   &   &  &$\checkmark$ &99.08  \\
			\hline
			9 & $\checkmark$   &    & $\checkmark$  & &5.5  \\
			\hline
			10 & $\checkmark$   & $\checkmark$   &   &  &98.11  \\
			\hline
			11 &  & $\checkmark$   & $\checkmark$  &$\checkmark$  &43.34  \\
			\hline
			12 & $\checkmark$   &   & $\checkmark$  &$\checkmark$  &98.98 \\
			\hline
			13 &$\checkmark$   & $\checkmark$   &   &$\checkmark$ &99.08 \\ 
			\hline
			14 & $\checkmark$   & $\checkmark$   & $\checkmark$  &  &97.57 \\ 
			\hline
			15 &$\checkmark$  &$\checkmark$   & $\checkmark$  &$\checkmark$  &\bm{$99.14$} \\ 
			\hline 
		\end{tabular}
		\caption{The performance evaluation of each loss function and their combination in HCRN.}
		\label{tab5}
	\end{table}
	$loss_{hpe}$ and $loss_{race}$. We can conclude that only the combination of $loss_{hpe}$ and $loss_{race}$ or adding extra loss of $loss_{kcls}$ performs best. Notice that the accuracy gets worse when we add the extra loss of $loss_{center}$.
	\item Row 15 in the table shows that when we use all the loss functions, HCRN achieves the best results compared with the previous cases, which also proves the importance of these four loss functions.
\end{itemize}

\subsection{Complexity Analysis}
In Tab.\ref{tab6}, we show the parameter, runtime memory, and speed of HCRN. The results are statistically calculated based on two experiments(unseen and seen characters). From col 2 we can see that HCRN consumes much memory on processing each character in both two experiments. The main reason for that is HCRN needs a lot of parameter calculation in the post-processing stage.

\begin{table}[htbp] \small
	\centering
	
	\begin{tabular}{|c|c|c|c|c|}
		\hline
		Classes & Mem & Time(GPU) & Time(CPU) & Para \\
		\hline
		27484 & $\sim$900MB & 10ms & 172ms &46.54MB \\
		\hline
		3755 & $\sim$123MB & 1ms & 25ms &46.54MB \\
		\hline
	\end{tabular}
	\caption{Investigation over parameter (Para), runtime memory on calculating each character (Mem), and speed (Time) of HCRN.}
	\label{tab6}
\end{table}

\section{Conclusions}
In this paper, we propose a novel method named HCRN to achieve smarter Chinese character recognition. One of the advantages of HCRN is that it can identify the whole characters by training partial radicals. To achieve this effect, we build a pseudo-siamese network to help the model to recognize unseen characters. Another advantage of HCRN is we don't need to label each structure information of characters.  We also evaluate the generalization performance of HCRN by identifying rotated characters. In the future, we plan to investigate HCRN's ability to recognize Chinese characters in natural scenes and look for the way to reduce the complexity of HCRN.

%\bibliographystyle{named}
\bibliographystyle{IEEEtran}
\bibliography{HCRN}

\end{document}